\documentclass[pdflatex,sn-mathphys-num]{sn-jnl}

\usepackage{graphicx}%
\usepackage{multirow}%
\usepackage{amsmath,amssymb,amsfonts}%
\usepackage{amsthm}%
\usepackage{mathrsfs}%
\usepackage[title]{appendix}%
\usepackage{xcolor}%
\usepackage{textcomp}%
\usepackage{manyfoot}%
\usepackage{booktabs}%
\usepackage{algorithm}%
\usepackage{algorithmicx}%
\usepackage{algpseudocode}%
\usepackage{listings}%

\begin{document}

\title[Article Title]{Transferable polychromatic optical encoder for neural networks}


\author*[1]{\fnm{Minho} \sur{Choi}}
\email{kernel@uw.edu}
\equalcont{These authors contributed equally to this work.}

\author[1]{\fnm{Jinlin} \sur{Xiang}}
\equalcont{These authors contributed equally to this work.}

\author[2]{\fnm{Anna}\sur{Wirth-Singh}}

\author[3]{\fnm{Seung-Hwan}\sur{Baek}}

\author*[1,4]{\fnm{Eli}\sur{Shlizerman}}
\email{shlizee@uw.edu}

\author*[1,2]{\fnm{Arka}\sur{Majumdar}}
\email{arka@uw.edu}

\affil[1]{\orgdiv{Department of Electrical and Computer Engineering}, \orgname{University of Washington}, \orgaddress{\city{Seattle}, \postcode{98103}, \state{WA}, \country{USA}}}

\affil[2]{\orgdiv{Department of Physics}, \orgname{University of Washington}, \orgaddress{\city{Seattle}, \postcode{98103}, \state{WA}, \country{USA}}}

\affil[3]{\orgdiv{Department of Computer Science and Engineering}, \orgname{Pohang University of Science and Technology}, \orgaddress{\city{Pohang}, \postcode{37673}, \state{Gyeongbuk}, \country{Republic of Korea}}}

\affil[4]{\orgdiv{Department of Applied Mathematics}, \orgname{University of Washington}, \orgaddress{\city{Seattle}, \postcode{98103}, \state{WA}, \country{USA}}}


\abstract{Artificial neural networks (ANNs) have fundamentally transformed the field of computer vision, providing unprecedented performance. However, these ANNs for image processing demand substantial computational resources, often hindering real-time operation. In this paper, we demonstrate an optical encoder that can perform convolution simultaneously in three color channels during the image capture, effectively implementing several initial convolutional layers of a ANN. Such an optical encoding results in $~24,000\times$ reduction in computational operations, with a state-of-the art classification accuracy ($\sim73.2\%$) in free-space optical system. In addition, our analog optical encoder, trained for CIFAR-10 data, can be transferred to the ImageNet subset, High-10, without any modifications, and still exhibits moderate accuracy. Our results evidence the potential of hybrid optical/digital computer vision system in which the optical frontend can pre-process an ambient scene to reduce the energy and latency of the whole computer vision system.}




\maketitle

\section{Introduction}\label{sec1}

Visual information plays a crucial role in human response, particularly in situations where reaction time is limited to a few tens to hundreds of milliseconds \cite{thorpe1996speed,graimann2010brain}. Though the human brain has efficiency far exceeding that of any other human-made computing systems, it still cannot process the entire collected visual data due to its massive amount of information. Most likely, our brain performs early visual processing to extract essential features for efficient and rapid interpretation without handling the entire visual data \cite{de2017understanding,iacaruso2017synaptic,nassi2009parallel}.

With the dramatic development of artificial intelligence (AI), computers can process the visual information like human brain, thanks to artificial neural network (ANN), enabling computer/machine vision \cite{yang2024vision,gehrig2024low,li2023intelligent,siemenn2024using, yang2024sensor}. Despite impressive progress, real-time inference with limited computational resources remains very challenging even with more efficient algorithms. For example, in a flying object (i.e., habitat drones \cite{woodget2017drones}) on-site data processing is plagued by severe heating, battery capacity and weight handling challenges. Utilizing cloud based systems poses challenges associated with data security and additional data transfer latency \cite{zhang2020vehicle,wu2023driver}.

Optical neural networks have emerged as a potential platform to circumvent these trade-offs, since an optical system can process multidimensional information with large spatio-temporal bandwidth \cite{mcmahon2023physics}. Recently, integrated photonics and free-space or fiber optics have been employed to implement some parts of an ANN for image compression/encryption \cite{wang2024integrated,chen2023photonic} and classification \cite{xue2024fully,bernstein2023single,xu2024large,xia2024nonlinear,bai2023all}. However, most of them are highly restricted on solving a relatively simple gray-scale datasets (i.e., MNIST and fashion-MNIST) and only a couple of systems have shown their implementation for more complicated multichannel datasets (i.e., CIFAR-10 and ImageNet) \cite{xue2024fully,xu2024large}. For these complex datasets, the optical system often become extremely large (with multiple stacks of the photonic circuits) \cite{xu2024large}, otherwise the classification accuracy remains low ($\sim60\%$ accuracy for CIFAR-10 classification tasks) \cite{xue2024fully,huo2023optical,rahman2023time}. In addition, the most successful ANN architectures utilize nonlinear activation functions that are challenging to implement optically. Proposed solutions, including atomic vapor cells \cite{Yang2023,Ryou2021} and image intensifiers \cite{wang2023image}, introduce significant experimental complexity, and additional power consumption.

To leverage the strengths of both optical and digital computing systems, an encoder-decoder inspired hybrid optical/digital architecture is a promising approach \cite{li2023intelligent,jang2024improved,yang2024sensor, ji2021cnn}. Specifically, an analog linear optical frontend (denoted as the optical encoder) performs bulk of linear computational tasks, while the digital backend implements the nonlinear operations. One intriguing possibility is to employ a static optical frontend, which is data agnostic, whereas the backend is trained and reconfigured. This resolves usual issues of modulation speed, errors, and system size in all-optical systems.  An optical encoder is particularly suitable for convolutional neural network (CNN) architectures, where convolutional layers act as feature extractors, encoding high-dimensional images into low-dimensional features \cite{xiang2022knowledge}. In fact, every free-space optic inherently performs a two-dimensional convolution operation during the imaging under incoherent light. The captured image is a convolution of the scene and the optic's incoherent point-spread-function (PSF) \cite{goodman2005introduction}. Thus, by engineering the PSF, an optical encoder can perform the desired convolution and replace the initial layers of a CNN.

Recently, PSF-engineered optical encoder has been employed to classify MNIST hand-written dataset and a reasonable classification accuracy with much less computational costs compared to the AlexNet is demonstrated \cite{wirth2024compressed}. We note that, however, MNIST images are monochrome, and is almost linearly separable ($0.84\space \%$ loss without any nonlinearity \cite{yang2010supervised}). The monochrome nature of the images makes the PSF-engineering approach wavelength agnostic. On the other hand, datasets such as CIFAR-10 \cite{alex2009learning} or ImageNet subset (High-10) \cite{coates2011analysis, xiang2022knowledge} are not separable by linear layers. Moreover, they consist of colored  images, where the actual color information is exploited in classification. 

Here, we demonstrate a polychromatic optical encoder with PSF-engineered meta-optics to classify the CIFAR-10 dataset. We first compressed the architecture into a single convolutional layer and two fully-connected layers using Knowledge Distillation. Then, we physically realized the convolution layer using an array of metasurfaces, where each metasurface, thanks to the inherent chromaticity, performs a separate convolution for each color channel. As a result, the hybrid CNN with an optical encoder reduces the total number of multiply–accumulate (MAC) operations at the digital backend by an factor of $\sim24,000$. The reduction of number of MAC operation directly corresponds to the computational costs, i.e., power and latency \cite{anderson2023optical}. It is worth noting that we always require an imaging system (i.e., lens and camera) to capture the image under ambient illumination, before we deliver the image data to the computational backend. Hence, with a single meta-optical encoder, we are not adding any additional optics, but simply replacing a conventional lens with PSF-engineered meta-optics. This makes our optical system compact and fully compatible with conventional optical imaging systems, while the other systems such as, integrated-photonic systems require pre-processing of the data\cite{xu2024large} and in-sensor computing needs a customized sensor design \cite{zhou2020near}. 

Furthermore, we adopt the same meta-optics (optical convolutional layer) which was optimized for CIFAR-10 dataset to High-10 dataset to explore the generality of optical encoders. In practice, a static optical encoder should be applicable for any scene. While, one approach is to employ reconfigurable frontend, e.g. based on non-volatile phase change materials \cite{tara2024non} or liquid crystals \cite{kang2024liquid}, the performance of these reconfigurable front end in terms of individual pixel control, power consumption, and operating speed are still inferior for practical deployment. Remarkably, with the same passive optical encoder (optimized for CIFAR-10 dataset), we achieved a high classification accuracy (for High-10 dataset) by fine-tuning the digital backend with additional fully-connected layer (via transfer learning approach). This ability to generalize the frontend is crucial for any ANNs as it enhances their versatility, efficiency, and robustness. A network that generalizes well can be applied to different tasks without extensive re-training, saving time, reducing costs for meta-surface fabrications, and conserving computational resources for real-world applications.

\section{Results}\label{sec2}

\begin{figure}[h]
\centering
\includegraphics[width=0.9\textwidth]{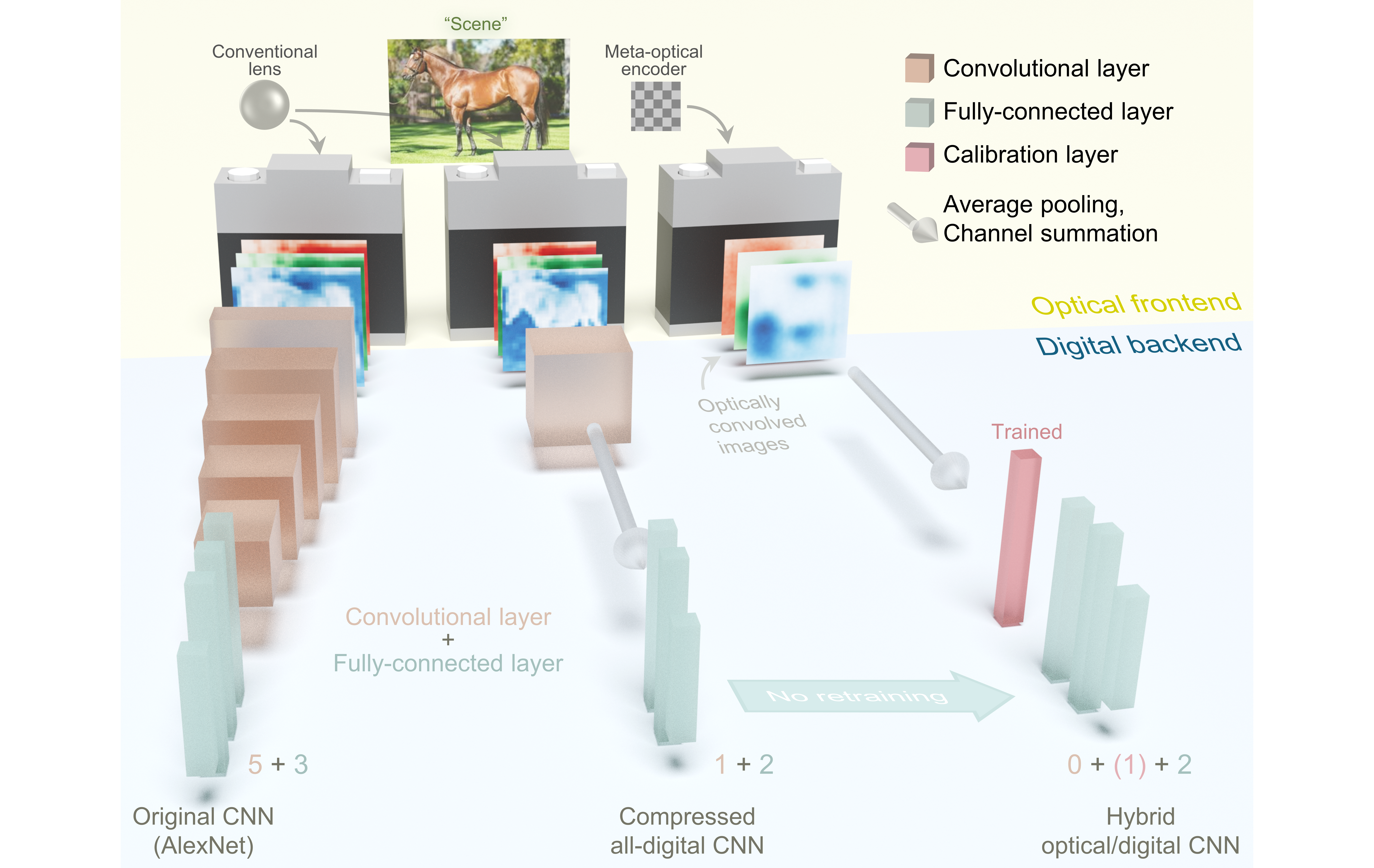}
\caption{\label{fig:Figure1}Schematic process flows of different image classification methods using an original CNN, a compressed all-digital CNN, and a hybrid optical/digital CNN.}
\end{figure}

Our optical encoder concept is described in Figure \ref{fig:Figure1}. The original CNN, i.e., AlexNet, has five convolutional layers and three max pooling layers at the front, followed by three fully-connected layers at the end, while nonlinear activation functions ``ReLU" are placed in each layer. Replacing all individual five convolutional layers with five sequential optics is extremely difficult because of misalignment, large system size, lack of nonlinearity, and low signal-to-noise ratio, issues that compound with increasing number of optical elements. Therefore, we compressed AlexNet to a single convolutional layer and two fully-connected layers using knowledge distillation method \cite{hinton2015distilling}, which reduces the complexity of the architecture with a minimal compromise in accuracy. 

While, the compression of an original CNN is essential for realizing the optical encoder scheme, there are practical trade-offs to consider.  On one hand, the physical size of the sensor and meta-optics  limit the number of kernels and kernel size. On the other hand, small kernel size or number of kernels fail to classify the data effectively. We empirically searched for the optimal number and size of the kernels while compressing the original CNN.  For CIFAR-10 classification task, we design 16 kernels of $7\times7$ size (see details in Supplementary Materials). The training and testing accuracy from the compressed all-digital CNN are $76.24\pm0.31\%$ and $75.90\pm0.30\%$, respectively. 

Since the CIFAR-10 images have three channel information $-$ corresponding to red (R), green (G), and blue (B) $-$ we have 16 individual $7\times7$ kernels for each color, a total of 48 kernels. In meta-optics, it is possible to design a single meta-optics to produce three different PSFs (i.e., convolutional kernels) for RGB wavelengths (more details in Methods). It is difficult to create positive and negative weights on the camera in terms of light intensity at the same time, so we separate each kernel into its positive and negative parts and design an optic for each \cite{chang2018hybrid}, for a total of 32 meta-optics in total corresponding to a 16 positive and negative polychromatic kernels.

Another important design parameter is to determine how many pixels on the camera represent one pixel of the PSF. We term this as the enlargement factor.  For example, when the enlargement factor is $2$, the ground-truth PSF which is a  $7\times7$ matrix will correspond to $14\times14$ pixels on the camera. While a large enlargement factor ensures less alignment error, the signal intensity on each camera pixel will be lower, resulting in low signal-to-noise ratio. To determine the optimal enlargement factor, we experimentally tested several meta-optics with different enlargement factors for a particular kernel (see details in Supplementary Materials), and obtained an optimal enlargement factor of $2$ for a meta-optic made of $3200\times 3200$ scatterers. 

\begin{figure}[h]

\centering
\includegraphics[width=0.9\textwidth]{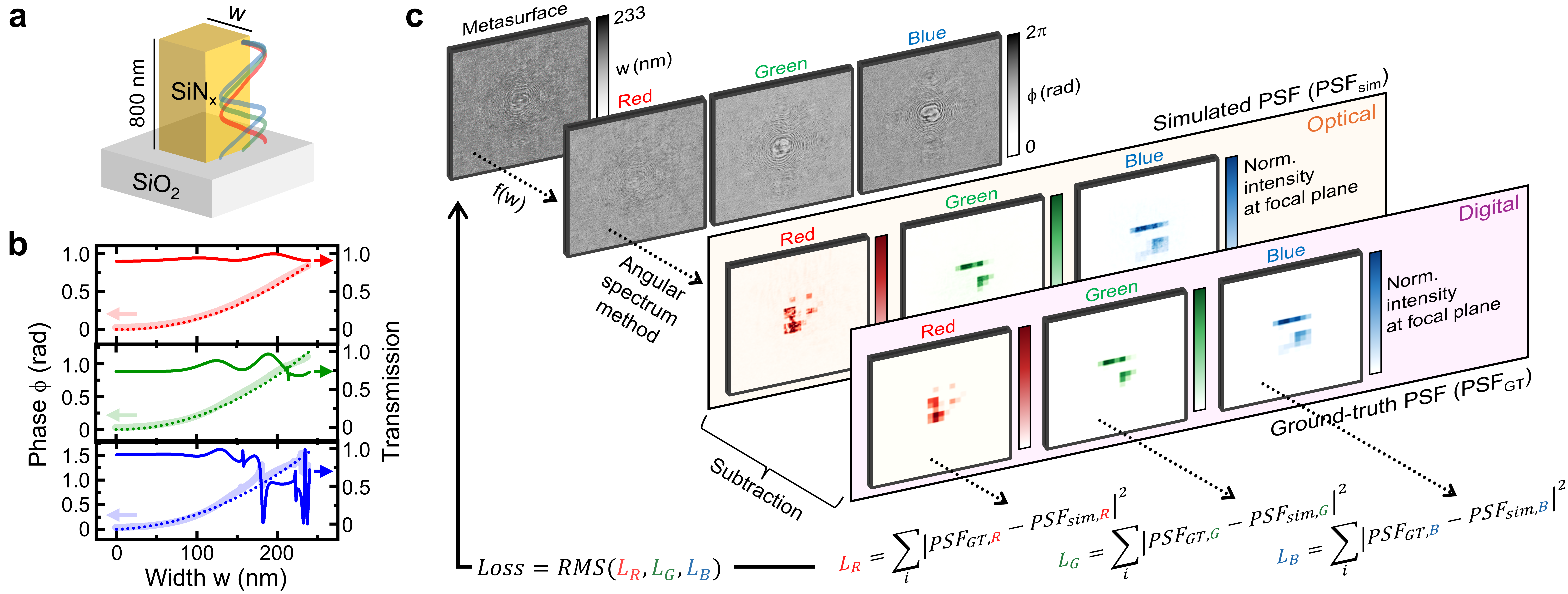}
\caption{\label{fig:Figure2} Polychromatic meta-optics design. (a) Schematics of the meta-optic scatterer, silicon nitride pillar on the quartz substrate. (b) Relative phase shift (dotted line) and transmission (solid line) of the uniform array of pillars with respect to the pillar width, $w$, for RGB different wavelengths when pillar height is fixed as 800 nm. Shaded line is the fitted proxy function of the phase shift with respect to the $w$. (c) Design flow of the polychromatic meta-optics optimization.}
\end{figure}

The metasurfaces were made of silicon nitride on a quartz substrate to ensure high transparency in the visible wavelength (Figure \ref{fig:Figure2}a). Figure \ref{fig:Figure2}b shows the transmission coefficients and phase shifts from silicon nitride pillars at RGB wavelengths as a function of the pillar width, $w$, at the fixed height of 800 nm, obtained by rigorous coupled-wave analysis (RCWA). We target wavelengths ($\lambda$) of $450$, $532$, and $635$ $nm$ for RGB colors based on the availability of the laser diodes.

In order to effectively model the wavelength-dependent effects of the meta-optics in a gradient descent-based optimization method, a fast and differentiable function is required to map between the pillar width and imparted phase. We define a proxy function inspired by the approximate phase shift of a dielectric waveguide with corrective factors that are fit to the RCWA simulation results. To calculate the phase shift ($f_R$, $f_G$, and $f_B$) for RGB wavelengths with respect to the pillar width, $w$, we define the proxy function as
\begin{equation} 
f_{\lambda}(w) = {2\pi n_{eff} L}/{\lambda} + A \exp{((w-B)^2/C)}-f_0. 
\end{equation}
The first term corresponds to a general phase shift from a dielectric waveguide, where $n_{eff}$ and $L$ are the effective refractive index and height of the silicon nitride pillars. The second term, only has the $w$ variance, corresponds to a correction term in the Gaussian shape with $A$, $B$, and $C$ as a fitting parameters. Lastly, $f_0$ corresponds to a phase shift offset, making the $f_\lambda(0)=0$. This proxy function does not model the resonance-induced phase variations; However, we do not want to use those phase variations due to reduced amplitude and these resonances are expected to be less prominent in the fabricated devices due to the sidewall roughness.

Figure \ref{fig:Figure2}c shows the design flow for the polychromatic RGB meta-optics that have optimized PSFs for individual RGB colors. For a meta-optic parameterized by an arbitrary two-dimensional pillar width map, $w(x,y)$, we extract three separate phase maps using the proxy functions $f_{\lambda}$. We then propagate the electromagnetic field using angular spectrum method \cite{shimobaba2012scaled} to simulate the PSFs at the focal plane, $2.4$ $mm$ away from the meta-optics. At the focal plane, we compare the  computational ground-truth PSFs defined by the convolutional kernels obtained using knowledge distillation ($PSF_{GT,\lambda}$) and optically-simulated PSFs ($PSF_{sim,\lambda}$) at each RGB channels, where the channel-dependent losses are defined by the sum of squares of differences in each pixels:
\begin{equation} 
\mathcal{L}_{\lambda} = \Sigma_{x,y} {\lvert PSF_{GT,\lambda}(x,y)-PSF_{sim,\lambda}(x,y)\rvert^2}. 
\end{equation}
We optimize the map of two-dimensional pillar width, i.e., meta-optics, for minimizing the net loss which defined as a root mean square of the losses at three different colors using the Adam optimizer in TensorFlow~\cite{developers2022tensorflow}:
\begin{equation} 
\min_{w(x,y)}{\lvert\lvert{\Sigma_{\lambda}{\mathcal{L}_{\lambda}^2}}\rvert\rvert^{1/2}}. 
\end{equation}
Calculated losses for all the kernels at three different colors are shown in the Supplementary Materials.

\begin{figure}[h]

\centering
\includegraphics[width=0.9\textwidth]{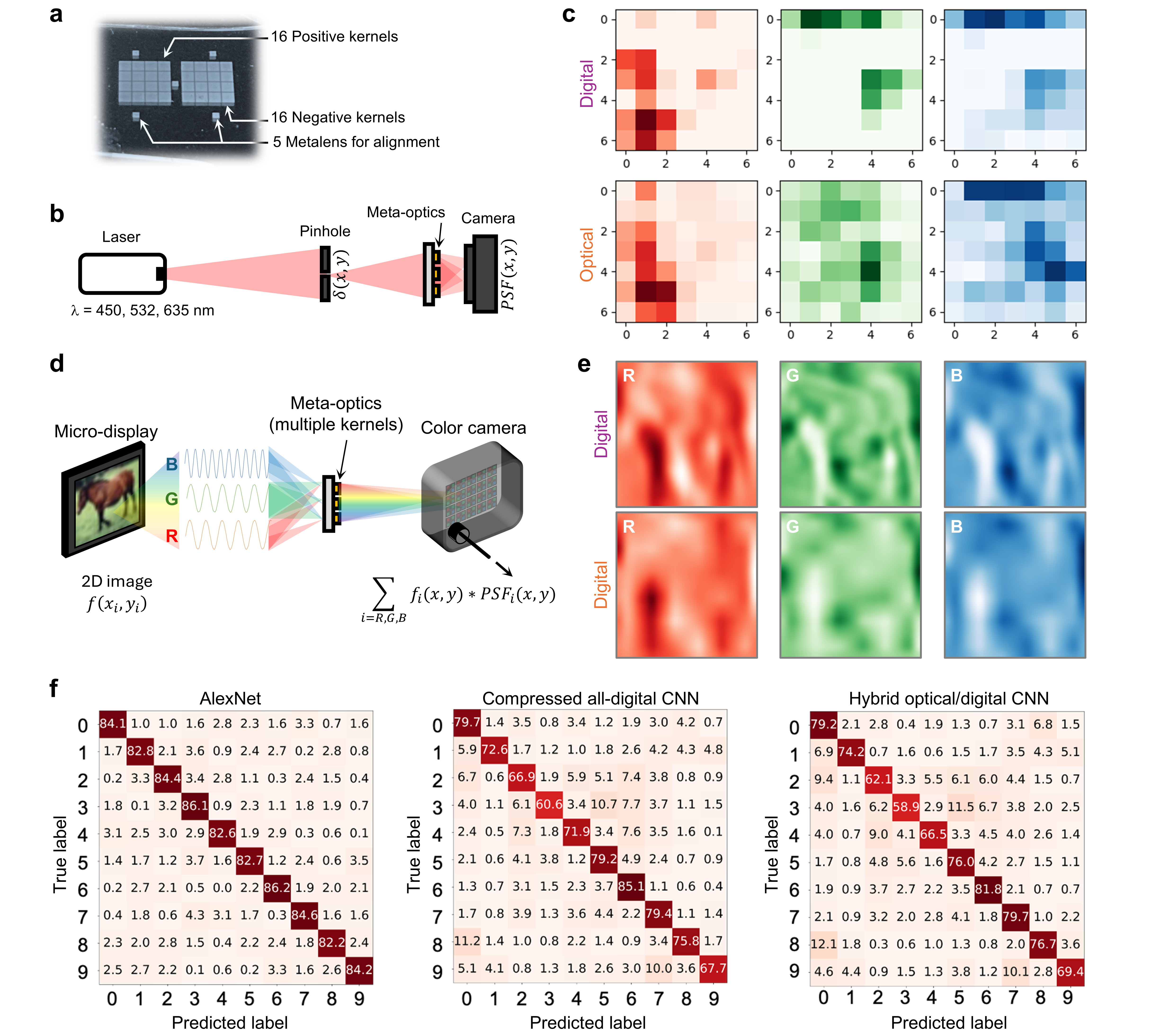}
\caption{\label{fig:Figure3} Optical characterizations of the meta-optic encoder. (a) Photograph of the fabricated optical encoder, consisting of 16 positive and 16 negative convolutional kernels and 5 algnment metalenses. (b) Schematics of the polychromatic PSFs measurement setup. (c) Ground-truth digital and optically measured PSFs for a particular polychromatic kernel (positive kernel number 7). (d) Schematics of the meta-optical convolved image measurement setup with a micro-display. A color camera capture convolved convolved images with a single shot. (e) Digital (above) and optical (below) convolution result of a particular CIFAR-10 image in individual RGB colors. (f) Confusion matrices of CIFAR-10 dataset classification tasks with different network architectures.}
\end{figure}

An optical image of the fabricated chip is shown in Figure \ref{fig:Figure3}a. A single chip contains a total of 32 convolutional meta-optics (corresponding to 16 positive and 16 negative convolutional kernels) and additional 5 metalenses which are focusing light at the focal plane, to aid in the alignment (e.g., tilt, rotation, and distance) between the meta-optics and the camera. Figure \ref{fig:Figure3}b shows the schematic of the PSF measurement setup. By changing the laser diodes, we illuminate individual RGB coherent light onto the camera through the meta-optics and experimentally characterize the polychromatic PSFs. A pinhole of $25 \space \mu m$ diameter creates an approximate point source and the positions of the optics, i.e., pinhole, meta-optics, and camera, remain the same while changing the laser diodes.

Figure \ref{fig:Figure3}c shows both the ground-truth PSFs and measured PSFs for a particular kernel for individual RGB wavelengths. To quantitatively analyze the difference between the ground-truth and experimentally measured PSFs, we define a cosine similarity ($\eta$) as:
\begin{equation} 
\eta = {\Sigma_i({A_iB_i})}/{\sqrt{\Sigma_i(A_i^2)}\sqrt{\Sigma_i(B_i^2)}}, 
\end{equation}
where $A_i$ and $B_i$ are the ground-truth and measured intensity profiles of the PSF for RGB wavelengths, respectively. The calculated $\eta$ for RGB wavelengths are about 0.88, 0.56, and 0.81, respectively. The quntitative discrepancy can be attributed partially to the fabrication and measurement imperfections. Additionaly, not all the polychromatic PSFs are physically realizable as the phases at different wavelengths are not completely independent. Creating more physically-realizable PSFs via co-designing the optical frontend and computational backend, also termed as end-to-end design \cite{huang2024photonic,wei2023spatially}, instead of replacing the convolutional layer with optics, may increase the $\eta$. However, including the meta--optical simulation in the end-to-end design may result in local optimum, and the fabrication/ measurement imperfections will still be present. As we will show later in the figure, the computational backend is robust against such discrepancy in the PSFs, and we can easily correct for these errors by introducing an additional fully-connected calibration layer in the digital backend.

Then, we test the polychromatic optical encoder for CIFAR-10 dataset. By replacing the pinhole with an organic light-emitting diode (OLED) display, we convolve the CIFAR-10 images with the characterized PSFs of the meta-optics (Figure \ref{fig:Figure3}d). The displayed image size is carefully adjusted according to the convolutional kernel size and the enlargement factor on the camera (more details in Supplementary Materials). Figure \ref{fig:Figure3}e shows the computationally and meta-optically convolved RGB images of one of the CIFAR-10 dataset. The meta-optically convolved image loses some of the high resolution components, likely due to the imperfect fabrication and alignment errors which are already recognisable from the PSF measurements as well as the spectral overlap between RGB color pixels of the camera. However, as we will show later in the figure, the computational backend is robust against such discrepancy as we do average pooling the convolved image into $6\times6$ size. We added an additional fully-connected layer, called calibration layer, to address the weights of each kernels and colors, dealing with the discrepancy between optical/digital systems (e.g., normalization, scaling, translation, rotation, tilt, noise). This calibration layer allows us to use the pre-trained digital backend and incur minimal computational cost. Detailed explanations on the calibration layer is in the Methods and Supplementary Materials.

\begin{table}[h]
\caption{Classification results on CIFAR-10 dataset}\label{table1}%
\begin{tabular*}{\textwidth}{@{\extracolsep\fill}ccc}
\toprule
Network Architecture & Train accuracy (\%)  & Test accuracy (\%)\\ 
\midrule
 AlexNet  & 83.04 $\pm$ 0.87  & 81.03 $\pm$ 0.89\\
Compressed digital CNN & 76.94 $\pm$ 0.52 & 76.59 $\pm$ 0.50\\
 Hybrid optical/digital CNN (w/o calibration layer) & 56.78 $\pm$ 0.91 & 56.39 $\pm$ 0.92\\
 Hybrid optical/digital CNN (w/ calibration layer) & 73.18 $\pm$ 0.58 & 72.06 $\pm$ 0.57\\ 
 Hybrid optical/digital CNN (w/) & 75.05 $\pm$ 0.47 & 73.17 $\pm$ 0.49\\
\botrule
\end{tabular*}
\end{table}

Figure \ref{fig:Figure3}f shows the confusion matrices of the classification accuracy of the CIFAR-10 data with an original CNN (AlexNet), a compressed CNN using knowledge distillation, and a hybrid optical/digital CNN using convolutional meta-optics after the compression. Even though there are slight differences between the optical and digital convolution results (Figure \ref{fig:Figure3}e), after introducing the calibration layer, we can achieve similar accuracy (less than 5\% loss) for both training and testing dataset (Table \ref{table1}). It is possible to improve the accuracy if we retrain the backend and the calibration layer; However, retraining the backend has no practical usages. Additionally, this hybrid approach significantly reduces computational costs which can be represented by the number of multiply-accumulate (MAC) operations. From the original CNN to compressed CNN, we can reduce the computational load, which is represented by a number of MAC operations, by a factor of $\sim 1,400$, while we can reduce further by a factor of $\sim 17$ after replacing a convolutional layer with meta-optics. Detailed calculation about number of MAC operations for each CNN architectures are described in Extended Data Table A1. The detailed information for network design and dimension selection is available in the Supplementary Materials.

\begin{figure}[h]
\centering
\includegraphics[width=0.9\textwidth]{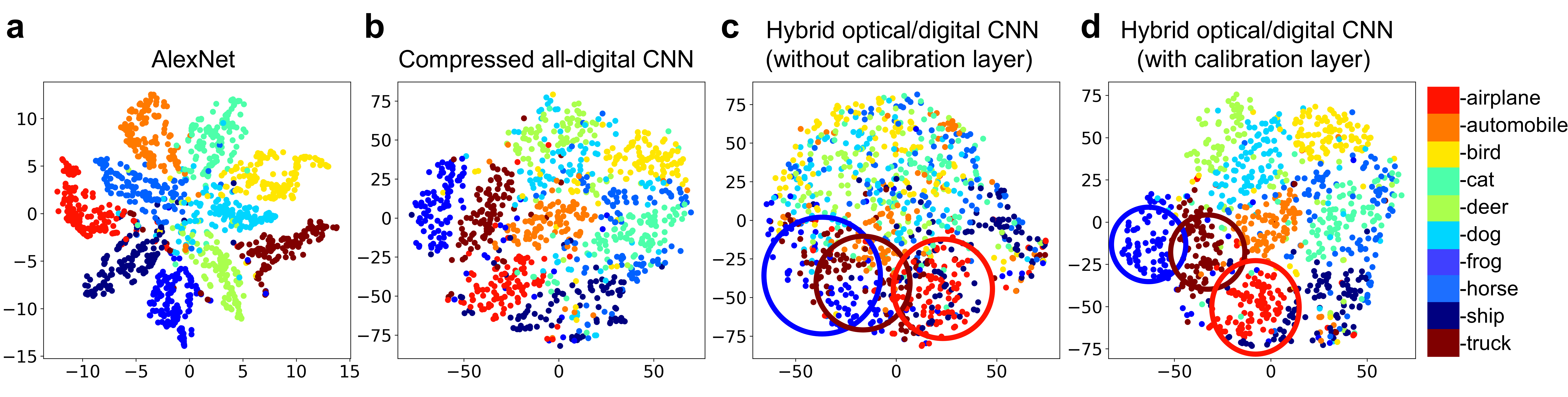}
\caption{\label{fig:Figure4} Principal component analysis for CIFAR-10 image dataset. (a) Original CNN, AlexNet. (b) Compressed all-digital CNN. (c) Hybrid optical/digital CNN without calibration layer. (d) Hybrid optical/digital CNN with additional calibration layer.}
\end{figure}

To analyze the effectiveness of the meta-optical convolutional layer, we utilize principal component analysis (Figure \ref{fig:Figure4}). For the original CNN and compressed all-digital CNN, each class is well-separated (Figures \ref{fig:Figure4} a and b), implying that we can extract out the key features of the CIFAR-10 image dataset after convolution. On the other hand, after the optical convolution using meta-optics, without calibration, different classes of the image were very difficult to distinguish (Figure \ref{fig:Figure4}c). Additionally, some clusters exhibit larger sizes and overlapping regions than Figure \ref{fig:Figure4}(a-b), e.g., the navy blue, brown, and red clusters (confidence ellipses). However, after introducing the calibration layer, the clustering regions become smaller, and the separations between classes increases. As shown in Figure \ref{fig:Figure4}d, each class becomes well-separated and distinguishable, similar to the compressed CNN. This critical role of the calibration layer is consistent with the classification accuracy without and with the calibration layer (Table \ref{table1}). We note that the calibration layer can potentially be compressed into the pretrained digital backend via additional training and will not affect the number of MAC operation for inference (Extended Data Table A1).

Our convolutional meta-optics implements convolutional kernels which came from compressed CNN for CIFAR-10 data. Unlike computational neural networks, optical implementations are extremely difficult to modify once they are fabricated. This necessitates different convolutional meta-optics for different dataset. However, we found that the convolutional layer that we optimized for CIFAR-10 can be readily adapted to classify another dataset High-10, with a transfer learning process. We added an additional fully-connected layer, which we call a ``transfer learning layer", that is located in between the former fully-connected layers and convolutional layer. By training the transfer learning layer, we can fit the other dataset, i.e., High-10, to the CNN which is pre-optimized for a particular dataset, i.e., CIFAR-10, with only fine-tuning a small part of the original network  without changing the former network structure (see details in Methods). The High-10 image dataset is polychromatic (RGB) and has a size of $224\times224$ size. To use the former CNN optimized for CIFAR-10 data for High-10 data, we resize the High-10 images to $32\times32$ size, same as the CIFAR-10 data. 

\begin{table}[h]
\caption{Transfer learning results on High10}\label{table2}%
\begin{tabular*}{\textwidth}{@{\extracolsep\fill}ccc}
\toprule
Network Architecture & Train accuracy (\%)  & Test accuracy (\%)\\ 
\midrule
AlexNet  & 85.31 $\pm$ 0.27  & 84.95 $\pm$ 049\\
Compressed CNN (without Transfer Learning) & 41.43 $\pm$ 0.26 & 40.44 $\pm$ 0.39\\
Compressed CNN (with Transfer Learning) & 67.43 $\pm$ 0.22 & 66.01 $\pm$ 0.13\\
Hybrid optical/digital CNN (with Transfer Learning) & 63.46 $\pm$ 0.46 & 59.73 $\pm$ 0.91\\
\botrule
\end{tabular*}
\end{table}

Without applying a transfer learning method, the training and testing accuracy is rather low around 40\%. However, after transfer learning, we achieve much higher training and testing accuracy ($\sim67.43\%$ and $\sim66.01\%$, respectively) on the High-10 data with the convolutional layer and two fully-connected layers. We further experimentally verified this approach works in our hybrid optical/digital CNN using the same convolutional meta-optics that we used for the CIFAR-10 data digital backend structure with one additional fully-connected layer. The average training and testing experiment accuracy of the High-10 data are similar (less than $5\%$ loss) to the compressed all-digital CNN, which is about the same of the CIFAR-10 case. The principal component analysis results of both compressed CNN and hybrid CNN for the High-10 data is shown in Supplementary Materials, where we can see how different classes are separable in their feature map. The detailed information for calibration layer design, number of calibration selection and PCA visualization is available in the supplementary materials.

\section{Discussion}\label{sec3}

\subsection{Multichannel dataset}\label{subsec1}

The advantages of the knowledge distillation and meta-optical encoder are a dramatic reduction of computational complexity, which is represented by the MAC operation. For the CIFAR-10 dataset, our hybrid optical/digital CNN reduced the number of MAC operation by a factor of $\sim24,000$. This reduction is about an order of magnitude higher than that of the MNIST hand-written dataset, where the meta-optical encoder reduced the number of MAC operation only by a factor of $\sim5,400$ \cite{wirth2024compressed}.

On the other hand, the classification accuracy drops are more significant for the CIFAR-10 dataset compared to the MNIST dataset. The train (test) accuracy for CIFAR-10 dataset of our hybrid CNN drops by $\sim9.86\%$ ($\sim8.97\%$) from the original CNN. For MNIST dataset, the train (test) accuracy of our hybrid CNN drops by $\sim5.0\%$ ($\sim5.0\%$) from the original CNN \cite{wirth2024compressed}. While this classification accuracy drop in CIFAR-10 dataset is not negligible, our PSF-engineered optical encoder has significantly large classification accuracy compared to the other free-space optical neural network architectures (which are compatible with conventional camera systems). Our encoder has a classification test (train) accuracy of $\sim73.2\%$ ($\sim72.1\%$) for CIFAR-10 dataset without retraining the backend and only projecting by a calibration layer. These test (train) accuracy can be improved further up to $\sim75.1\%$ ($\sim73.2\%$) if we retrain the backend, which is better than the previous state-of-the-art result ($\sim72.8\%$) which used a complex end-to-end optimization as well as the backend retraining with 50 number of kernels \cite{wei2023spatially}. Our hybrid optical/digital CNN can be further improved by using complex meta-atoms to reproduce better PSFs optically \cite{Huang24broadband} and using advanced compression method to reduce the loss during the knowledge distillation \cite{kim2016sequence}. The other reports have much less accuracy $\sim63\%$\cite{xue2024fully,huo2023optical,rahman2023time} compared to ours. 

For a ImageNet subset, High-10, we have the same amount of reduction in number of MAC operation as the CIFAR-10 dataset since we used the identical CNN architecture. The train (test) accuracy of our hybrid CNN heavily drops by $\sim21.85\%$ ($\sim25.22\%$) compared the original CNN. The majority of losses occur during the network compression as we share the convolutional layer and fully-connected layers optimized for CIFAR-10 dataset. However, albeit to the losses, our transfer learning results has a classification accuracy of $\sim61\%$, still better than the other free-space optical neural networks system \cite{xue2024fully}. Here, we selected the ImageNet dataset which has more complicated and distinct classes from the CIFAR-10 dataset. We did not change the optical frontend, but only fine-tuned the digital backend of two fully-connected layers and an additional transfer learning layer, to show the versatility of our hybrid CNN system.

\subsection{Energy consumption}\label{subsec2}
In practice, we can implement our hybrid optical/digital CNN simply by replacing a lens with meta-optics during imaging. Hence, the energy consumption will solely be determined by the number of MAC operations. However, it is important to also consider the power from the sensor.  Specifically the sensor power depends on the number of pixels being passed to the digital backend. For an original CNN, we only need $32\times32$ pixels to capture the image. On the other hand, hybrid CNN needs $6\times6$ pixels for imaging one convolved image, considering the average pooling (details in the Supplementary), which ends up with $32\times6\times6$ pixels for all positive and negative kernels. Hence, our hybrid CNN requires a bit larger number of pixels for imaging compared to the original CNN.

The color camera we used (Allied Vision Prosilica; GT 1930 C) has a total power consumption of $3.4W$ with $50.70$ frames per second and $1,936\times1,216$ pixels, which ends up with $28nJ$ per frame and pixel. Thus we estimate that the original CNN and hybrid CNN requires an energy of about $29.1\mu J$ and $32.8\mu J$, respectively, for the image capturing process per a single image. However, the energy consumption for the computational backend is much larger for the original CNN compared to the hybrid CNN. For state-of-the-art computational system, an energy consumption per a single MAC operation is $\sim1pJ$\cite{wang2023image,huang2024photonic}. Thus the energy consumption for a single object classification task for the hybrid CNN is about $150nJ$, which is more than four orders of magnitude smaller than that of the original CNN, $3.65mJ$. We note that, we can trade-off the sensor power (by reducing the number of kernels) with computational backend power (by increasing MAC operations). However, having more operation in the optical encoder with a simple computational backend will always be preferred to reduce the latency.

\subsection{Applications}\label{subsec3}
The hybrid CNN has a strong advantages in terms of latency and energy consumption compared to the original CNN. Additionally, it can be adequately integrated on the commercial imaging system (e.g., camera) without modifying the physical architecture other than the lens with a PSF-engineered meta-optics. Moreover, the ability to encode a colorful image brings up a potential to utilize the encoder for real-world scenes. However, sacrifice of the accuracy is extremely crucial and not negotiable for cases where the safety matters (e.g., autonomous driving vehicles). In other words, if the object classification is applied for statistical analysis (where the ensemble average can minimize the individual inaccuracy), we can endure the loss of classification accuracy. Habitat monitoring drones can be an example \cite{koger2023quantifying,chalmers2021video}. Especially, in case of drones, restrictions to minimize their weight is crucial, which force it store only essential features. Then on-site data processing can be beneficial. As our optical encoder can minimize the latency and energy consumption for the on-site data processing, the habitat drones can investigate much larger areas with a single flight.

\section{Conclusion}\label{sec4}
The results in this work are strong evidence that optical frontend can significantly reduce the power consumption and latency of ANNs for computer vision tasks. Despite realistic fabrication and measurement errors arising from optical implementation, the approach achieves the state-of-the art classification accuracy in multichannel CIFAR-10 data with the addition of a calibration layer and trainable fully-connected layers.  The use of a single meta-optical layer to perform complex, multi-channel convolutions highlights a unique applicability of meta-optics that cannot be accomplished using traditional optics. In addition, we address the lack of reconfigurability in existing optical implementations using transfer learning approach, and reconcile the optical frontend optimized for CIFAR-10 to the High-10 dataset. In this regard, we suggest that a hybrid approach comprised of an optical frontend and reconfigurable digital backend utilizes the key advantages of optics (i.e., no latency, no loss, large space-bandwidth) with robustness and reconfigurability provided by the backend. 

\section{Methods}\label{sec5}

\subsection{Knowledge distillation}\label{subsec5}
Typically, the knowledge distillation algorithm is designed to compress neural networks. Here, we propose using knowledge distillation to transfer the generalized knowledge from a larger, pre-trained teacher network, AlexNet, to a more compact CNN, referred to as the ``student network." Specifically, the student network comprises only a single convolutional layer coupled with a backend, which consists of a single fully connected layer and a linear calibration layer. In addition, we selected AlexNet as our teacher network for two main reasons: first, AlexNet was the foundational model that successfully addressed the ImageNet dataset. Additionally, compared to more complex networks like ResNet-18 or VGG-16, the five-layer AlexNet is more accessible and easier to implement optically.

The knowledge distillation algorithm includes two types of losses: student loss and temperature loss. Student loss minimizes the discrepancy between the student network's predictions and the ground truth labels. The softmax function is used to compute:

\begin{equation} 
p_{i}^{\text{student}} = \frac{\exp(z_i)}{\sum \exp(z_i)} 
\end{equation}

where $z_i$ represents the student logits after the last fully connected layer. Temperature loss, on the other hand, optimizes the discrepancy between the student network's predictions and the teacher network's predictions. Knowledge distillation incorporates a softening parameter, $T$, known as the distillation temperature for the teacher probabilities. Thus, we can compute such loss as:
\begin{equation} 
p_{i}^{\text{temperature}} = \frac{\exp(y_i/T)}{\sum \exp(y_i/T)} 
\end{equation}
Finally, the total loss is calculated as a weighted sum of the two losses:
\begin{equation} 
\mathcal{L}(x, \Phi) = \alpha \mathcal{L}_C(y, p^{\text{student}}) + (1 - \alpha) \mathcal{L}_K \left( p^{\text{temperature}}; T = \tau, p^{\text{student}}; T = \tau \right) \label{eq: loss} 
\end{equation}
where $\alpha$ is the weight balancing the two loss components, $\mathcal{L}_C$ is the cross-entropy loss function, $\mathcal{L}_k$ is the Kullback-Leibler (KL) Divergence loss function~\cite{seo2020kl}.

We also find that other key hyperparameters might impact our hybrid CNN system. First, in most CNNs, such as ResNet-18 and AlexNet, there are multiple convolutional layers, each with more than 200 kernels to extract useful features and maintain generalization across various datasets. While some pruning strategies show that using 1\% of the parameters can achieve similar accuracy~\cite{liang2021pruning}, applying these algorithms to optical neural networks are non-trivial. Most pruning methods still retain ANN structures, which suffer from misalignments that are nearly impossible to eliminate~\cite{wirth2024compressed}. Therefore, compressing into shallower layers with more kernels is preferred. However, each meta-surface has a physical size that limits the number of kernels it can contain. To address this limitation, we can use multiple cameras and multiple meta-surfaces to increase the number of kernels, thereby improving the classification accuracy and generalization of the hybrid CNN.

\subsection{Meta-optics design}\label{subsec6}
For 16 digital kernels for each R, G, and B channels, we have 32 meta-optical kernels as we use a single meta-optics for all RGB channels but we cannot represent both positive and negative weights with optics. Hence, we create 16 positive kernels and 16 negative kernels, then perform digital substraction on the digital backend. Each of our convolutional meta-optics has $3200\times3200$ scatterers, with $2\times2$ scatters constitute a group to enhance the robustness of fabrication. Based on the ground-truth digital convolutional kernels, we defined optical PSFs for each RGB channels, and inverse-design the meta-optics having those PSFs at each RGB wavelengths using TensorFlow Adam optimizers.

\subsection{Meta-optics fabrication}\label{subsec7}
Our meta-optics operate at visible wavelength ($\lambda \sim400$ $nm-700$ $nm$). We use silicon nitride on quartz substrate for the meta-optics to have high transparency at the whole visible regime. We deposit a thick silicon nitride layer ($800$ $nm$) on top of the double-polished quartz substrate using plasma-enhanced chemical vapor deposition (Oxford; Plasma Lab 100). We spin coat and bake electron beam resist (ZEP-520A) on top of the silicon nitride layer, followed by a spin coat anti-charging agent (DisCharge H20). We pattern using electron beam lithography (JEOL; JBX6300FS), and develop the resist using amyl acetate. After that, we deposit via electron beam evaporation (CHA; SEC-600) and do lift-off an alumina layer ($\sim65$ $nm$) for hard mask . Finally, we etch the silicon nitride with an alumina hard mask using plasma etcher with fluorine-based gas (Oxford; PlasmaLab 100, ICP-180). The sub-wavelength structured meta-optics has a period of 293 nm, which is a half of the camera pixel size collecting the image. 

\subsection{Optical measurements}\label{subsec8}
We measure the PSF by placing a laser and a pinhole ($\phi=25$ $\mu m$), representing a point source. Then we place the convolutional meta-optics on 3-axis stage with rotational knobs to align the meta-optics centered and parallel to the beam path. High resolution color camera (GT-1930C) which has a pixel size of 5.86 $\mu m$ is placed $2.4\space mm$ away from the meta-optics. We measure the PSFs for each RGB color light by replacing the laser with three different wavelengths (Thorlabs; CPS450, CPS532, and CPS635). For image convolution measurements of the CIFAR-10 dataset, we put the micro-display at the pinhole position, then connect to the computer to show the color images. Since a single meta-optics can represent three different RGB kernels at the same time (see details in Supplementary Materials), a color camera which have RGB color pixels can extract the convolved images at three different channels. This can eventually save the space of the meta-optics and camera, which is critical in real-world applications. The point source is replaced by an arbitrary two-dimensional image, $f(x,y)$. We can express the image as a sum of the three color channels, $f_R(x,y)+f_G(x,y)+f_B(x,y)$. The convolutional meta-optics perform a convolution for each color, and as a result, a convolved image, $\sum\limits_{i=R,G,B}{f_i(x,y)\ast PSF_i(x,y)}$, will be imaged on the camera. Since we determined the enlargement factor of $2$ for the PSF, we use the same enlargement factor for the CIFAR-10 image as well. According to the camera pixel size, $5.86$ $\mu m$, and and CIFAR-10 image size, $32\times32$, the projected image size on the camera has to be about $374$ $\mu m\times374$ $\mu m$. At the given values of distance between the display and meta-optics and meta-optics to the camera, we can end up with the CIFAR-10 image size on the display to be $16.0$ $mm\times16.0$ $mm$. We use 10,000 images for training (subset of original $50,000$ images )and 10,000 images for testing, with an exposure time of 500 ms. Among the 10,000 images of training and testing dataset, 186 and 201 images are not involved on training and testing, respectively, due to the overexposure issue. All the measurement parameters and number of images are the same for the High-10 dataset for transfer learning process.

Another critical factor is the exposure time. Since the optical features are captured by a CCD camera, the exposure time significantly influences the final performance. If the optical features are overexposed, texture information, such as the fur of a cat, might be missing. Conversely, if the optical features are underexposed, most information may also be lost, resulting in a lack of distinction between highlights and shadows in the image. To find the most appropriate exposure, we could use a similar approach to modern cameras, where ``18\% gray" is considered as the mid-point between black and white on a logarithmic or exponential curve. This standard can help us achieve balanced exposure, ensuring that the captured optical features are neither overexposed nor underexposed.

\subsection{Computational backend}\label{subsec9}
As previously discussed, optical fabrication and alignment noise are unavoidable in meta-surface kernels. These include scaling, translation, rotation, image aberration, and optical noises. To address this issue, we propose adding a calibration function to remap the optical convolution outputs to align with those of the previously trained backend. Specifically, we use a fully connected layer as the calibration function and corresponding loss function is defined as:

\begin{equation}
\mathcal{L} = \min(f_{\text{calibrate}}(\textit{Hybrid optical/digital CNN}, \textit{Compressed CNN}))
\end{equation}

This approach aims to refine the experimental outputs to align more closely with the pre-designed network. To prevent overfitting, we strategically limit our training to only 20\% of the available data, ensuring that our model remains efficient~\cite{xiang2022tkil，xiang2023tkil,zheng2020bi}.

\subsection{Transfer learning}\label{subsec10}
Generalization is a key feature to test our hybrid optical/digital CNN. Ensuring that the network can generalize well to new, unseen data is crucial for several reasons. First, our hybrid network is compressed from AlexNet, which was originally designed with a large dataset. The pre-trained AlexNet achieves high accuracy across various datasets and can be easily adapted or fine-tuned to out-of-distribution datasets. This adaptability is essential for practical applications where the data distribution may differ from the training set. Second, exploring the generalization capabilities of hybrid models is important because designing and fabricating different meta-surface kernels for different tasks is inefficient. By enhancing generalization, we can use a single hybrid model for multiple tasks, reducing the need for extensive redesigns and fabrications. Details and schematics of our transfer learning plan are shown in the Supplementary Materials.

To implement the transfer learning, we add two types of losses: feature loss and label loss. The feature loss minimizes the discrepancy between the optical features and the digital features, ensuring that the representations learned by the optical and digital components are aligned. The label loss minimizes the discrepancy between the model’s predictions and the actual labels, improving the overall prediction accuracy. During the transfer learning process, the optical frontend and digital backend remain unchanged. We add two fully connected layers between the optical front end and backends and fine-tune these layers using the two losses. Specifically, the function is:

\begin{equation}
\mathcal{L} = \alpha \mathcal{L}_{\text{feature}} + \beta \mathcal{L}_{\text{label}}
\end{equation}

where $\mathcal{L}_{\text{feature}}$ is the feature loss and $\mathcal{L}_{\text{label}}$ is the label loss, with $\alpha, \beta$ as the respective weights balancing these losses.

\section*{Declarations}\label{sec6}

\backmatter

\bmhead{Acknowledgements}
The research is supported by National Science Foundation (EFRI-BRAID-2223495).  Part of this work was conducted at the Washington Nanofabrication Facility/ Molecular Analysis Facility, a National Nanotechnology Coordinated Infrastructure (NNCI) site at the University of Washington with partial support from the National Science Foundation via awards NNCI-1542101 and NNCI-2025489.

\bmhead{Conflict of interest}
A.M. is a co-founder of Tunoptix, which aims to commercialize meta-optics technology.

\bmhead{Code availability}
The code and data generated and/or analyzed are available from the corresponding author upon reasonable request.

\bmhead{Author contribution}
M.C. and J.X. contributed equally to this work. A.M. and E.S. conceived of the idea and provided funding. A.W.S. involved in developing the PSF-engineering approach. J.X. and E.S. developed the knowledge distillation approach. J.X. trained the neural network and analyzed the experiment data. M.C. developed the polychromatic PSF-engineering approach and designed the meta-optics. M.C. fabricated the meta-optics and conducted the optical experiments. S.H.B. advised on the result analysis. M.C., J.X., and A.M. wrote the manuscript with input from all authors. 

\begin{appendices}

\section{Extended Data}\label{secA1}
\begin{table}[h]
\caption{Number of MAC operations depend on CNN architectures (*can be compressed)}\label{table3}
\begin{tabular*}{\textwidth}{@{\extracolsep\fill}cccc}
\toprule
Layers&AlexNet &Compressed CNN &Hybrid optical/digital CNN\\ 
\midrule
&3$\times$64$\times$11$\times$11$\times$224$\times$224&&\\
&64$\times$192$\times$5$\times$5$\times$55$\times$55&&\\
Convolution&192$\times$384$\times$5$\times$5$\times$27$\times$27&3$\times$16$\times$7$\times$7$\times$32$\times$32&0\\
&384$\times$256$\times$3$\times$3$\times$13$\times$13&&\\
&256$\times$256$\times$3$\times$3$\times$6$\times$6&&\\
\midrule
&9216$\times$4096&576$\times$256&576$\times$576$^*$\\
Fully-connected&4096$\times$1024&256$\times$10&576$\times$256\\
&1024$\times$10&&256$\times$10\\
\midrule
Sum&3,651,368,960&2,558,464&150,016\\
\botrule
\end{tabular*}
\end{table}

\section{Supplementary Materials}\label{secA2}
The online version contains supplementary materials.




\end{appendices}


\bibliography{sn-bibliography}

\end{document}